\pdfoutput=1
\documentclass{article}

\usepackage[preprint]{neurips_2024}

\usepackage{fancyhdr}
\usepackage{nth}
\usepackage[utf8]{inputenc} % allow utf-8 input
\usepackage[T1]{fontenc}    % use 8-bit T1 fonts
\usepackage{hyperref}       % hyperlinks
\usepackage{url}            % simple URL typesetting
\usepackage{booktabs}       % professional-quality tables
\usepackage{amsfonts}       % blackboard math symbols
\usepackage{nicefrac}       % compact symbols for 1/2, etc.
\usepackage{microtype}      % microtypography
\usepackage{xcolor}         % colors
\usepackage{graphicx}
\usepackage{hyperref}
\usepackage{amsmath}
\usepackage{cleveref}
\usepackage{color, colortbl}
\usepackage{tabularray}
\usepackage{multirow}
\usepackage{nicematrix}
\usepackage{caption}
\usepackage{subcaption}
\usepackage{booktabs} % For better table formatting
\usepackage{wrapfig}

\definecolor{mygray}{gray}{0.9}

\title{2COOOL: 2nd Workshop on the Challenge Of Out-Of-Label Hazards in Autonomous Driving}

\definecolor{mydarkblue}{rgb}{0,0.08,1}
\definecolor{mydarkgreen}{rgb}{0.02,0.6,0.02}
\definecolor{darkred}{rgb}{0.8,0.02,0.02}
\definecolor{darkorange}{rgb}{0.40,0.2,0.02}
\definecolor{darkpurple}{RGB}{111,0,255}
\definecolor{myred}{rgb}{1.0,0.0,0.0}
\definecolor{mygold}{rgb}{0.75,0.6,0.12}
\definecolor{mydarkgray}{rgb}{0.66, 0.66, 0.66}

\author{%
   \parbox{\textwidth}{\centering
   Ali K. AlShami$^{1,\dagger}$, Ryan Rabinowitz$^{1}$, Maged Shoman$^{2}$, Jianwu Fang$^{3}$, Lukas Picek$^{4}$ \\
   \textbf{Shao-Yuan Lo$^{5}$, Steve Cruz$^{6}$, Khang Nhut Lam$^{7}$, Nachiket Kamod$^{1}$, Lei-Lei Li$^{3}$ }\\
   \textbf{Jugal Kalita$^{1}$, Terrance E. Boult$^{1}$} \\[6pt]
   $^{1}$University of Colorado Colorado Springs, 
   $^{2}$University of Tennessee--Oak Ridge Innovation Institute,  \quad
   $^{3}$Xi'an Jiaotong University, \quad 
   $^{4}$University of West Bohemia \quad
   $^{5}$Honda Research Institute USA \quad
   $^{6}$University of Notre Dame \quad
   $^{7}$Can Tho University \\[6pt]
   \texttt{aalshami@uccs.edu}, \texttt{rrabinow@uccs.edu}, \texttt{mshoman@utk.edu}, \texttt{j.w.fangit@gmail.com}, \\
   \texttt{lukaspicek@gmail.com}, \texttt{shao-yuan\_lo@honda-ri.com}, \texttt{stevecruz@nd.edu}, \texttt{klam2@uccs.edu},
   \texttt{nachikamod@gmail.com}, \texttt{670160532lileilei@gmail.com}, \texttt{jkalita@uccs.edu}, \texttt{tboult@vast.uccs.edu}}
}

\begin{document}

\maketitle

\begin{abstract}
As the Computer Vision community advances autonomous driving algorithms, integrating vision-based insights with sensor data remains essential for improving perception, decision-making, planning, prediction, simulation, and control. Yet we must ask: It’s 2025—why don’t we have entirely safe self-driving cars yet? A key part of the answer lies in addressing novel scenarios, one of the most critical barriers to real-world deployment. Our 2COOOL workshop provides a dedicated forum for researchers and industry experts to push the state-of-the-art in novelty handling, including out-of-distribution hazard detection, vision–language models for hazard understanding, new benchmarking and methodologies, and safe autonomous driving practices. The “2nd Workshop on the Challenge of Out-of-Label Hazards in Autonomous Driving” (2COOOL) will be held at the International Conference on Computer Vision (ICCV) 2025 in Honolulu, Hawaii, on October 19, 2025. We aim to inspire the development of new algorithms and systems for hazard avoidance, drawing on ideas from anomaly detection, open-set recognition, open-vocabulary modeling, domain adaptation, and related fields. Building on the success of its inaugural edition at the Winter Conference on Applications of Computer Vision (WACV) 2025, the workshop will feature a dynamic mix of academic and industry participation.
\end{abstract}

\section{Introduction}
\label{sec:introduction}
The deployment of autonomous vehicles is rapidly expanding, driven by advances in AI, machine learning, and sensor technology that promise to improve safety and mobility. However, ensuring safe operation in open-ended, real-world environments requires that autonomous systems not only recognize known entities but also reliably anticipate and respond to novel situations. Current self-driving systems perform well at detecting predefined objects (e.g., cars, pedestrians) under familiar conditions, but they often struggle with anomalous or unexpected obstacles outside their training labels~\cite{alshami2024coool}. This limitation can leave the vehicle “blind” to hazards it was never trained to see, undermining reliability.

\begin{figure}[h] % 'h' means place it here
    \centering
    \includegraphics[width=0.8\textwidth]{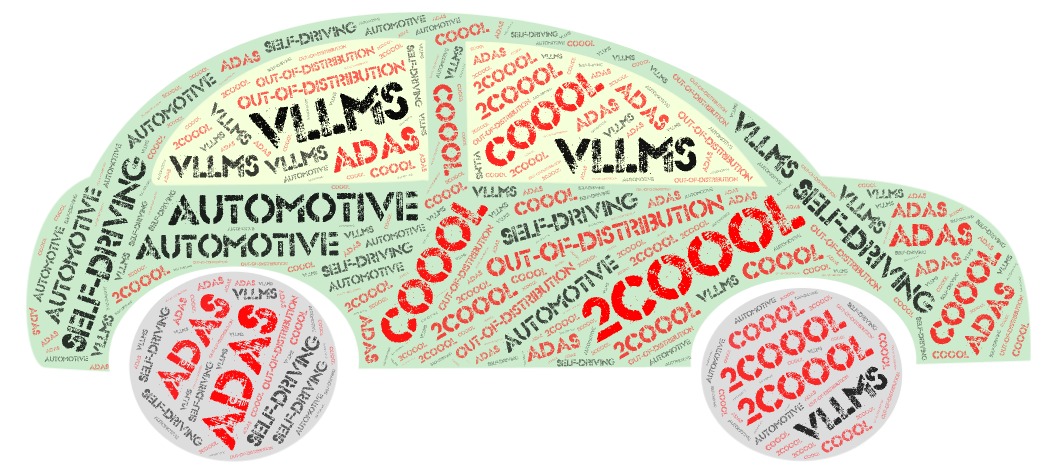} % Adjust width and filename
    \caption{Visual embodiment of 2COOOL interests.}
    \label{fig:my_label}
\end{figure}

This novelty problem exposes autonomous vehicles to scenarios for which they lack prior knowledge, elevating the risk of catastrophic outcomes if such hazards are not handled properly. Achieving true robustness, therefore, requires the ability to detect and appropriately react to novel dangers beyond the standard object ontology. Out-of-distribution hazards are inherently difficult to learn because they are by definition not present in training data and can be very expensive to collect from deployed systems. Traditional autonomous driving datasets like KITTI \cite{geiger2012we} and nuScenes \cite {caesar2020nuscenes} have been instrumental for object detection and tracking, providing large-scale labeled data across many modalities. However, these benchmarks focus on a fixed set of object classes and do not explicitly evaluate a system’s ability to handle novel or anomalous obstacles. Indeed, a recent survey found \cite{liu2024survey} that even though some driving datasets encompass dozens of object categories, they still lack the diverse and rare examples necessary to assess performance on out-of-distribution events.

In safety-critical applications, the capability to estimate uncertainty and detect failures is as important as achieving high accuracy on nominal cases \cite{ghiya2025sgnetpose+}. For example, the Fishyscapes benchmark was introduced to measure anomaly detection in urban semantic segmentation, and it revealed that identifying unexpected objects (via model uncertainty) is far from solved, even in ordinary city street scenes \cite{blum2021fishyscapes}. In short, most autonomous vehicles today operate under a closed-world assumption, the expectation that test-time inputs will resemble the training distribution. Bridging the gap to an open-world setting, where new hazards can appear at any time, remains an urgent research challenge.

Researchers have begun tackling the novelty challenge through multiple avenues. In the field of open-set recognition (OSR), algorithms are designed to recognize when an input does not belong to any known class \cite{shrivastava2023novelty}. The OSR problem was first formalized as requiring classifiers to not only label known classes correctly but also reject inputs from novel, unknown classes beyond their training categories \cite{scheirer2012toward, alshami2024smart}. Alongside classification-based OSR, uncertainty-driven methods currently being explored include: energy-based confidence scoring and generative (autoregressive) models, which aim to better distinguish in-distribution versus out-of-distribution objects in images. Despite these efforts, detecting novelty is a continuum and safe deployment of real-world systems remains a critical area of intense research.

In the autonomous driving domain specifically, several works have integrated novelty detection mechanisms into the driving stack \cite{du2022unknown}. For example, Amini et al. (2018) incorporated a variational autoencoder into an end-to-end driving model to detect when the vehicle encounters an unusual scenario. This approach allowed the system to flag novel observations by recognizing patterns it had never seen during training. Similarly, Greer and Trivedi (2024) explored the use of language embeddings in an active learning framework to identify novel scenes in driving data \cite{greer2024towards}. By leveraging a vision-language model to describe or embed scenes, their method could detect out-of-distribution events and then actively query for human annotations, thereby gathering new examples of rare events. These works illustrate a growing recognition that robust autonomous driving requires more than mastering a fixed training set – the vehicle must be prepared to handle the open-world variability of the real road.

In parallel, the advent of large-scale foundation models is opening promising new directions for open-world driving safety. Vision and multimodal foundation models, trained on diverse, Internet-scale datasets, can endow autonomous systems with a broader understanding of visual context \cite{alshami2023pose2trajectory}. Such models have demonstrated a remarkable ability to capture a wide range of visual concepts and distributions in the world~\cite{liu2024can}. By exploiting this general knowledge, using for example, through open-vocabulary object detectors or by synthesizing rare hazard scenarios, an autonomous vehicle may better recognize out-of-label objects with minimal task-specific training. This approach aligns closely with the mission of our workshop: leveraging generalizable AI (spanning vision, language, and learning from vast data) to help foresee and safely handle the “unknown unknowns” on the road.

To advance research in this area, the Challenge of Out-of-Label (COOOL) benchmark was recently proposed as one of the first datasets dedicated to novelty in driving. COOOL comprises a collection of over 200 high-resolution dashcam videos with meticulous annotations of both common and uncommon road hazards. Importantly, the dataset captures many rare obstacles that are typically absent from training data – including exotic animals on the roadway (e.g., kangaroos, wild boars) and inanimate debris or other anomalies (e.g. plastic bags blowing across the road, smoke on the roadway) – alongside the usual cars, bicycles, pedestrians, and other standard objects. Each video is exhaustively labeled frame-by-frame, and a unique tagging system records the driver’s reactions and vehicle maneuvers, providing rich context on how human drivers respond to these surprise hazards. By evaluating algorithms on their ability to detect unknown hazards and predict potential dangers in these real-world clips, the COOOL benchmark aims to fill a critical gap in testing open-world driving safety. It provides a much-needed testbed for measuring how well new models can generalize to out-of-label events that were never seen during training.

The 2COOOL workshop at ICCV 2025 builds on these developments, bringing together the community to address out-of-label hazards from multiple angles \cite{2coool2025}. In the following sections, we outline the scope of the workshop and its key topics, summarize the accepted papers (grouped by theme), and introduce the invited experts who will share their insights on novelty and safety in autonomous driving.

\section{Workshop Scope and Topics}
\label{sec:related_work}

The central aim of 2COOOL is to broaden the discussion of autonomous driving safety beyond the conventional, labeled scenarios. The workshop scope covers detection, prediction, and mitigation of hazards that current self-driving systems are not explicitly trained to handle. This includes a range of research topics and open problems:
\begin{itemize}
    \item \textbf{Novel Hazard Detection and Recognition:} \\
    How can an AI driver detect and recognize an object or situation that was never seen in training? This spans anomaly detection in images, open-set object detection, and open-vocabulary recognition in the driving context. For instance, detecting a new obstacle type (like an animal on the highway) or recognizing an off-nominal traffic event are problems of interest. Advances in this area draw on techniques from one-class classification and open-set recognition to equip models with a “reject option” for unknowns.
    \item \textbf{Hazard Prediction in Video:} \\
    Detecting static images of unknown hazards is only part of the challenge – the workshop also highlights temporal understanding, i.e., predicting hazards in a sequence of frames. This involves activity understanding and forecasting: given a video, which objects or agents are likely to become hazardous in the next moments? Solutions here intersect with trajectory forecasting and intention prediction, needing to anticipate rare events (e.g., a pedestrian about to jaywalk or an animal leaping onto the road).
    \item \textbf{Handling Low-Resolution or Hard-to-See Obstacles:} \\
    Some hazards are inherently difficult to perceive – for example, distant objects or objects in poor lighting. The workshop topics include methods for dealing with low-resolution hazards, pushing the limits of sensor processing and super-resolution to ensure even faint signs of danger are caught early.
    \item \textbf{New Datasets and Benchmarks:}  \\
    Because novel hazards are by definition rare, creating datasets that capture these events is crucial. The community is encouraged to propose new datasets, simulation environments, and evaluation metrics that specifically target out-of-label scenarios. By sharing data on “corner cases” and establishing benchmarks like COOOL, researchers can systematically measure progress on this problem.
    
    \item \textbf{Vision and Vision-Language Models for Open-World Driving:} \\
    With the rise of large-scale vision-language models, there is interest in how they can contribute to safer driving. For example, a multimodal model might use broad semantic knowledge (from web-scale training) to identify an unusual object on the road even if the specific instance was never labeled in driving datasets. The workshop welcomes work on applying foundation models and open-vocabulary detectors to driving, bridging the gap between general AI and domain-specific safety.

    \item \textbf{Human Factors and Explainability:} \\
    Finally, ensuring safety isn’t just about the perception algorithms – it’s also about how the system’s decisions are understood and trusted by humans. The scope includes human factors (HRI) in autonomous driving and explainable AI (XAI) for hazard handling. For example, if an autonomous car slows down for an “unrecognized object,” how can it explain this decision to the safety driver or passenger? And how do we design interfaces or alerts for human takeover in the face of novel dangers?
\end{itemize} 

Collectively, these themes underscore the workshop’s message: improving autonomous vehicle safety requires going beyond the closed set of training labels and tackling the open-world challenges of novelty, uncertainty, and rare events. By emphasizing out-of-label hazards, 2COOOL shines a spotlight on the “long tail” of driving scenarios that are easy to overlook but vital for truly robust autonomy.
\section{Accepted Papers}
\label{sec:dataset}
We received a strong set of submissions from researchers and practitioners addressing the challenges of novelty in autonomous driving. After a rigorous peer-review process, we accepted 12 high-quality papers covering a wide range of topics aligned with the workshop themes. These papers span areas such as:

\begin{itemize}
    \item VRU-Accident: A Vision-Language Benchmark for Video Question Answering an Dense Captioning for Accident Scene Understanding.
    \item Uncertainty-Aware Likelihood Ratio Estimation for Pixel-Wise Out-of Distribution Detection.
    \item Adapt, But Don’t Forget: Fine-Tuning and Contrastive Routing for Lane Detection under Distribution Shift.
    \item Interpretable Decision-Making for End-to-End Autonomous Driving.
    \item Fourier Domain Adaptation for Traffic Light Detection in Adverse Weather.
    \item DRAMA-X: A Fine-grained Intent Prediction and Risk Reasoning Benchmark For Driving.
    \item SADWA: Fine-Grained Weather Awareness with Vision-Language Models for Seamless Autonomous Driving in Real Time.
    \item FlareGS: 4D Flare Removal using Gaussian Splatting for Urban Scenes.
    \item Towards Vision Zero: The Accid3nD Dataset.
    \item Simplifying Traffic Anomaly Detection with Video Foundation Models
    \item Ctrl-Crash: Controllable Diffusion for Realistic Car Crashes
    \item LaViPlan: Language-Guided Visual Path Planning with RLVR

\end{itemize}

The full list of accepted papers, along with their presentation formats (oral/poster), will be made available on the \href{www.2coool.net}{2COOOL website} and in the ICCV 2025 workshop proceedings. 

\section{Invited Speakers}
\label{sec:approach}
To spark meaningful dialogue, 2COOOL features invited talks from leading experts in computer vision and autonomous driving—spanning academia and industry—whose work strongly aligns with the workshop’s focus on novelty and safety. Below, we present concise bios of the invited speakers and highlight how their expertise addresses out-of-label hazard challenges.

\begin{itemize}
  \item \textbf{Dr. Walter J. Scheirer (University of Notre Dame)} \\
  \textbf{Bio:} A leading voice in open‑set recognition and anomaly detection. His work on statistical models like W‑SVM and EVM formalizes the “open‑world” problem, emphasizing models that recognize when inputs fall outside known classes.
 
  \item \textbf{Dr.\ Manmohan Chandraker (UC San Diego \& NEC Labs America)}
  \textbf{Bio:} His research interests are in vision, learning, and graphics, with applications to autonomous driving and augmented reality. His works have been recognized with best paper awards or honorable mentions at CVPR, ICCV, and ECCV, the NSF CAREER Award, Qualcomm, and Google Research Awards. He serves on NSF panels on vision, learning, and robotics and on senior program committees at CVPR, ICCV, ECCV, AAAI, ICML, NeurIPS, and ICLR.

  \item \textbf{Dr.\ Xiatian Zhu (University of Surrey, CVSSP)} \\
  \textbf{Bio:}  a Senior Lecturer affiliated with the Surrey Institute for People-Centred Artificial Intelligence and the Centre for Vision, Speech and Signal Processing (CVSSP) at the University of Surrey in Guildford, UK, leads the Universal Perception (UP) Lab. Previously a research scientist at Samsung AI Centre, Cambridge, UK, Dr. Zhu holds a Ph.D. from the Queen Mary University of London.

  \item \textbf{Dr.\ Zhengzhong Tu (Texas A\&M University)} \\
  \textbf{Bio:} Dr. Tu, a former Research Engineer at Google Research, specializes in AI and computer vision, with 30+ publications in top venues like TPAMI, CVPR, and NeurIPS. He co-organized workshops at ITSC, WACV, and CVPR and won the AI4Streaming 2024 Challenge. His work has earned the CVPR 2022 Best Paper Nomination and media recognition from Google Research, YOLOvX, and Hugging Face.
  
  \item \textbf{Dr.\ Hesham Eraqi (Amazon)} \\
  \textbf{Bio:} In 2018, he became Valeo Senior Expert/Director of AI, leading Comfort and Driving Assistance Systems. At Valeo (2012–2021), he contributed to the Honda Legend, the first road-legal Level 3 autonomous car (Japan, 2021). He was an Adjunct Assistant Professor at AUC, holds six patents and 40+ publications, and serves on Program Committees for top AI and autonomous driving conferences.

  \item \textbf{Mr.\ Paul E. (CEO, 3LC.AI)} \\
  \textbf{Bio:} Paul Endresen is a technology entrepreneur and innovator with over 30 years of experience. He co-founded Innerloop Studios, creating hit titles such as Joint Strike Fighter, SEGA Extreme Sports, and Project IGI, and later co-founded Hue AS, leading the development of its industry-agnostic cloud platform. As CTO of Bluware, he pioneered InteractivAI, a real-time AI copilot for image interpretation. His work revealed the critical importance of data quality in machine learning, inspiring him to found 3LC.

  \item \textbf{Dr. Nemanja Djuric (Aurora)} \\
  \textbf{Bio:} Dr. Nemanja Djuric is a Principal Technical Lead Manager at Aurora Innovation, where he has played a key role over the past four and a half years in leading the development and deployment of the perception system behind the world’s first commercially-launched fully autonomous truck (early May 2025). Prior to joining Aurora, Nemanja spent nearly six years at Uber’s Advanced Technologies Group, driving innovation in perception and forecasting for self-driving vehicles. Earlier in his career, he was a Research Scientist at Yahoo Labs, where he focused on computational advertising. His contributions to the field earned him recognition in 2022 as one of the Top 100 Global Researchers in Information Retrieval and Recommendation, as listed by ArnetMiner.

  \item \textbf{Dr. Hamed Tabkhi (University of North Carolina)} \\
  \textbf{Bio:} Tabkhi is an Associate Professor of Electrical and Computer Engineering at the University of North Carolina at Charlotte. His research focuses on developing novel computer vision algorithms and system architectures to address real-world challenges, particularly in transportation safety, public safety, and public health. Dr. Tabkhi emphasizes close collaboration with domain experts and community stakeholders to ensure meaningful and applicable solutions. The National Science Foundation recognized his Smart and Connected Communities project as a program success story.

  \item \textbf{Dr. Shubham Shrivastava (Kodiak Robotics)} \\
  \textbf{Bio:} Shubham heads AI and machine learning at Kodiak, pushing the limits of generative AI, vision-language models, large foundation models, and spatio-temporal multimodal networks. By fusing camera, lidar, radar, and language signals, his group delivers a holistic, time-consistent 3-D understanding of the world that steers every autonomous decision.
  \item \textbf{Roni Goldshmidt from Nexar} \\
  \textbf{Bio:}
  AI researcher focused on deep learning perception models for autonomous systems, specializing in foundation models, video-to-video generation, and vision-language models at Nexar. Co-founder of SimpML and active open-source contributor, with published research advancing AI for safer, smarter autonomy.
\end{itemize}

This diverse set of keynote speakers exemplifies the workshop’s dual emphasis: advancing academic research while addressing industry needs. Academic speakers will delve into the latest algorithms and theories for novelty detection and safety, whereas industry speakers will highlight real-world considerations (such as system reliability, engineering constraints, and product impact). By hearing from both sides, workshop attendees gain a comprehensive understanding of the out-of-label hazard problem. The keynotes are expected to stimulate discussion on how cutting-edge research can be translated into deployable systems, and conversely, how practical challenges can inspire new academic inquiries. We are proud to host these talks and look forward to the knowledge exchange they will foster
\section{Organizers' experience and background}
\label{sec:experiments}

\begin{itemize}
    \item \textbf{Ali K. AlShami} has over ten years of experience in academia and industry, specializing in AI and ML/CV. He has contributed to multiple funded research projects and has published in, as well as served as a peer reviewer for, prestigious conferences and journals. Ali is the main chair of the \href{https://sites.google.com/ucr.edu/cooolsworkshop/home}{COOOL} and \href{https://2coool.net/}{2COOOL} workshops at WACV 2025 and ICCV 2025.

    \item \textbf{Ryan Rabinowitz} is a PhD candidate at the University of Colorado, Colorado Springs, where he has participated in numerous research efforts supporting DARPA, IARPA, NGA, and the U.S. Army. He co-organized the first \href{https://sites.google.com/ucr.edu/cooolsworkshop/home}{COOOL} and second \href{https://2coool.net/}{2COOOL} workshops at WACV 25 and implemented and managed the Kaggle competitions. He has published multiple peer-reviewed articles focused on open-set recognition and incremental learning, including in ICCV 25, AAAI 25, and ECCV 2024.
    
    \item \textbf{Maged Shoman} is a researcher in Deep Learning, Computer Vision, and Autonomous Vehicles, with work supported by NSF and U.S. DOT. He has 10+ publications in venues like IEEE CVPR and ASCE Journal of Transportation Engineering and has received multiple awards, including the NVIDIA AI City Challenge.

    \item \textbf{Jianwu Fang} His research focuses on an intelligent understanding of driving scenarios. He has 90+ publications, 11 patents, and developed a large-scale (14M+ frames) traffic accident dataset. He has earned top awards, including the Shaanxi Natural Science Award, and serves as an Associate Editor for IEEE T-ITS. 
    \item \textbf{Lukáš Picek} specializes in applied machine learning for species recognition, distribution modeling, animal identification, and tram autonomous driving. Since 2021, he has co-organized CVPR-FGVC and LifeCLEF workshops and leads the CVPR tutorial on Animal Re-identification. As FGVC competition chair, he has organized multiple Kaggle competitions with thousands of participants.

    \item \textbf{Steve Cruz} received his Ph.D from the University of Colorado Colorado Springs. His research focuses on open-set and open-world learning problems, novelty detection, the integration of domain knowledge in meta-learning, and hyperparameter optimization.
    
    \item \textbf{Khang Lam} earned a Ph.D. in Computer Science from UCCS in 2015. Her research focuses on Machine Learning, Computer Vision, NLP, and Image Captioning. She has experience collecting video data for autonomous driving systems and has organized workshops in Asia.
    \item \textbf{Jugal Kalita} research focuses on NLP, Computer Vision, and Machine Learning, with applications in biology and security. He has supervised over 125 undergraduates, 100+ MS students, and 25 Ph.D. students. Jugal also organized some workshops at prestigious NLP conferences. 
    
    \item \textbf{Terrance E. Boult} has a 40-year academic history and is an IEEE Fellow, past IEEE PAMI TC Chair, Co-founder, and treasurer of the Computer Vision Foundation.  He has been a general,  program or chair of CVPR 4 times, general or program chair of WACV 5 times, and a finance or other chair dozens of times.
\end{itemize}
\section{The 2COOOL Hazard-Report Challenge}
\label{sec:compitition}

Autonomous driving systems generate enormous data streams, while out-of-label hazards are rare and crucial to improve system design and training.
Our ICCV workshop competition, the second Challenge Of Out-Of-Label (2COOOL), addresses this problem by tasking participants to build systems capable of sifting through normal driving data to find valuable hazard and accident data, and generate incident reports.
The 2COOOL dataset builds upon the original COOOL benchmark, now integrating three distinct datasets, including COOOL \cite{alshami2024coool} DADA \cite{fang2021dada}, and Nexar \cite{moura2025nexar}. By combining these resources, 2COOOL covers a wide spectrum of driving situations, from everyday conditions to unusual and unexpected road events.

\begin{figure*}[h]
    \centering
    \includegraphics[width=1\textwidth,height=0.4\textheight]{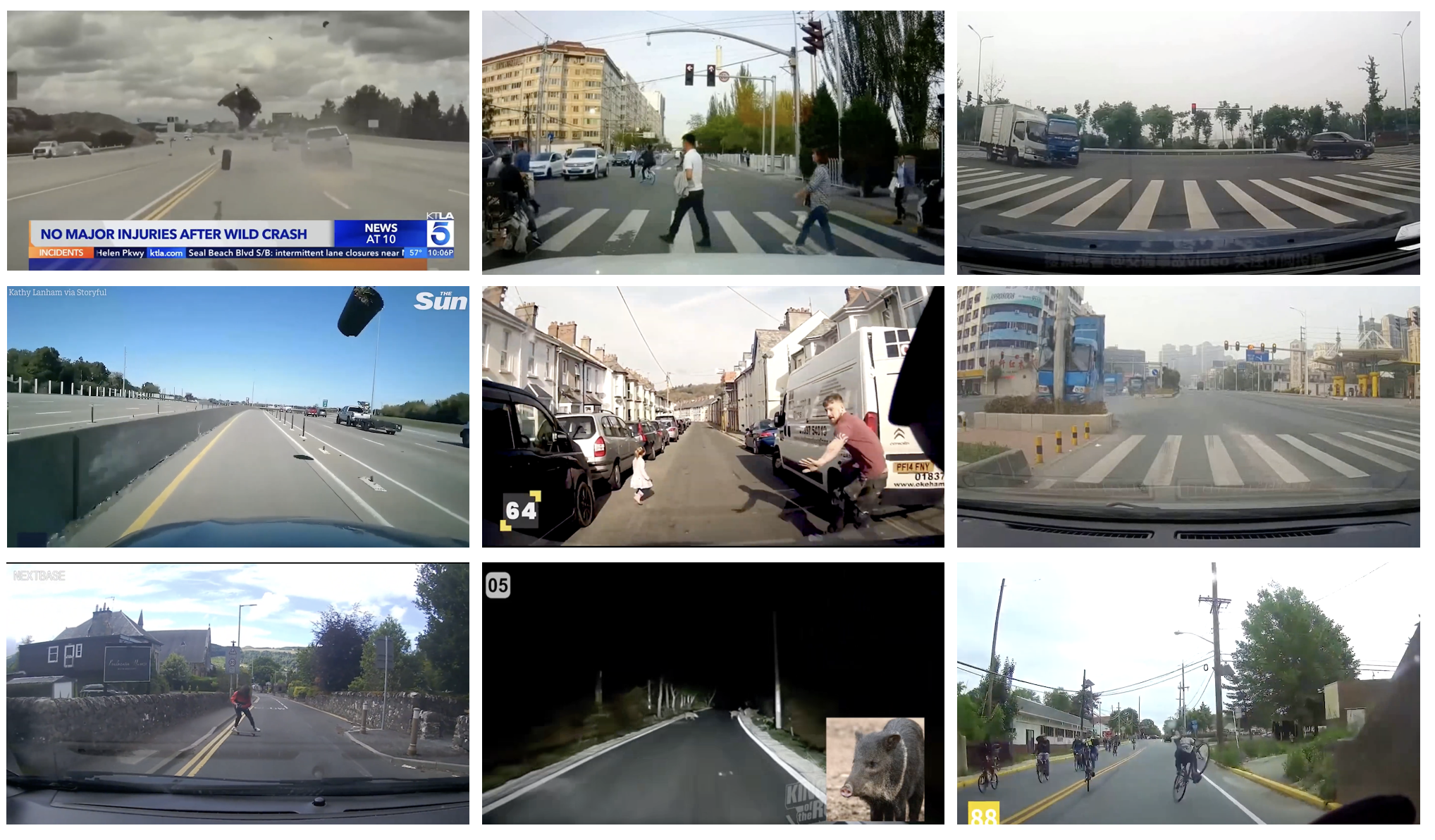}
    \caption{Examples from the 2COOOL Benchmark.}
    \label{fig:2coool_examples}
\end{figure*}

In the first COOOL challenge (WACV 2025) \cite{coool2025}, participants tackled three core tasks: (i) detecting driver reactions (identifying if/when the driver reacts to a hazard), (ii) hazard object identification, and (iii) hazard captioning. Notably, the inclusion of hazard captioning in that inaugural competition demonstrated the value of describing hazards in natural language, complementing traditional detection metrics. Building on this insight, the new 2COOOL challenge places greater emphasis on the vision-and-language aspect. Participants are now tasked with producing coherent textual descriptions of the hazards and incidents captured in dashcam footage, rather than just detecting them.

The updated 2COOOL dataset provides rich contextual information around each incident. It includes detailed captions describing the moments before and after each event, along with annotations indicating whether the event involves a hazard, an accident, or no accident at all. Additional labels specify crucial details such as: whether the scene is hazardous or not, the severity of any crash, whether the ego-vehicle (the camera car) is involved, counts of other parties involved (vehicles, pedestrians, cyclists, animals, etc.), the time-to-hazard (how far ahead the hazard is, in frames or seconds), and even driver gaze information with corresponding gaze captions. These annotations were created by multiple sources – in fact, four human annotation teams and multiple Vision-Language Models (with human verification) contributed captions – to ensure a diverse and comprehensive set of ground-truth descriptions.

Instead of naively asking a VLLM to caption an entire video (which would be inefficient and prone to irrelevant details), the competition breaks the problem into key sub-tasks. By first extracting essential information from the video, the final report generation becomes more focused and accurate. The 2COOOL challenge provides participants with critical information upfront, structured into five tasks, which serve as building blocks for the final hazard report:
\begin{itemize}
    \item Time-to-Hazard/Accident Estimation (time2hazard): Predict how much time (or how many frames) remain until a hazard or accident occurs, effectively estimating the moment of the incident in the video.
    \item Hazard/Accident Detection: Determine whether the scene contains a hazardous situation, an actual accident, or no significant incident. (In other words, classify each video clip as “hazard,” “accident,” or “no incident.”)
    \item Hazard/Accident Recognition: If there is a hazard or accident, identify the type or category of it. For example, is the hazard a jaywalking pedestrian? Debris on the road? A vehicle running a red light? This task assigns a label describing the nature of the hazard or accident.
    \item Ego-Car and Other Party Involvement: Ascertain whether the ego-vehicle (the one with the dashcam) is involved in the hazard/accident. Additionally, report the involvement of other parties by category and count (e.g., how many other cars, pedestrians, cyclists, or animals are part of the event).
    \item Crash Severity Assessment: If an accident or crash occurs, assess its severity or level of danger from a list of defined levels.
    
\end{itemize}
Participants will generate predictions for these five aspects for each video. The outputs from these sub-tasks are then fed into a VLLM to produce a detailed, context-rich caption that describes the hazard or accident. In essence, the system first figures out what happened, when, and who/what was involved, and then uses that information to articulate a concise narrative of the event. To support this approach, the competition organizers have compiled and annotated approximately 700 dashcam videos as the challenge dataset. Each video has been annotated with ground-truth labels for the above tasks, and multiple reference captions were written by four human annotation teams and multiple large VLLMs (with human verification). This provides a strong basis for both training the models and for fairly evaluating the generated reports (since multiple reference descriptions help capture the varied ways a hazard can be described).

Challenge Task: Using the above information, each participating team’s system must automatically generate a hazard or accident report for every provided dashcam clip (no report for no-incident clip). The report should be concise yet informative, accurately describing the critical hazard(s) or incident in the scene. It needs to convey the nature of the hazard, relevant context, and even the outcome or a possible recommended action in a human-understandable way. For example, a report might say “a deer suddenly darts across the road, forcing the driver to brake abruptly” or “debris falls from a truck ahead, creating an obstacle and causing the following car to swerve.” The description should effectively communicate what went wrong or what potential danger is present, as if you were explaining the incident to someone who wasn’t there. Achieving this is quite challenging — it requires integrating video understanding (to recognize events, objects, and actions in the footage) with natural language generation (to clearly and correctly articulate those observations). In other words, the system has to both “see” what’s happening and “explain” it in words.

Motivation: By requiring textual explanations of driving incidents, the 2COOOL hazard-report task pushes beyond simple yes/no or category outputs and moves toward explainable AI for autonomous driving. Rather than just detecting that a hazard exists, the system must explain what the hazard is and why it is dangerous, a valuable task when working with huge amounts of data. Recent research underscores the importance of this capability. For instance, when the original COOOL dataset was extended with natural-language hazard annotations, it enabled a much richer evaluation of anomaly detection methods, the models weren’t only judged on if they detected an anomaly, but also how well they could describe the anomaly. Descriptive outputs can capture subtle details about why a scenario is dangerous, details that purely numeric scores or category labels might miss. Moreover, considering the temporal sequence of events is crucial in understanding many hazards: a single snapshot might not show the full context of an emerging danger. A situation often becomes hazardous due to the way something unfolds over time (for example, noticing a pedestrian gradually veering into the road, or an initially distant object suddenly accelerating into the vehicle’s path). By using video clips and requiring a narrative report, the 2COOOL challenge forces participants to perform such temporal reasoning. Systems need to connect the dots across frames, e.g. recognizing an object’s motion that will intersect with the ego-car’s trajectory, or detecting a delayed reaction from the driver, to truly understand and explain the hazard. This focus on dynamic, time-based understanding reflects real driving needs and encourages the development of models that handle the dynamic nature of driving scenes, rather than treating each frame in isolation.

Evaluation: Submissions will be evaluated on both the accuracy and the clarity of their generated reports. We plan to use a combination of automated metrics and human judgment to assess the quality of the outputs. 
%For automated evaluation, one approach is to measure the semantic similarity between a generated report and the ground-truth description of the same incident. This could involve embedding-based similarity scores or even evaluations by another language model, essentially checking if the AI captured the correct gist of the hazard, even if it uses different wording. 
We will incorporate traditional text generation metrics for participant feedback like SPICE, CIDEr, METEOR to gauge overlap with reference texts, but given the open-ended nature of this task, such metrics are not sufficient alone. It’s important that the reports are written in fluent English and include the most relevant details of the incident. Therefore, automated checks will be complemented by expert human review for the top submissions, where incident reports will be ranked. Human judges will examine whether the descriptions are accurate, clear, and useful (for example, would a traffic safety analyst find the report informative and correct?). The full evaluation protocol, including which metrics will be used and how, and providing sample ground-truth reports, will be detailed on the Kaggle competition page. This will ensure participants know what kind of output format and content is expected, and how their submissions will be scored.

Timeline and Participation: The 2COOOL hazard-report competition is slated to open on Kaggle in the near future (exact dates to be announced on the competition page). When it launches, participants around the world will receive the training dataset of annotated dashcam videos and example hazard reports. They will have a period of several weeks to develop and train their models on this data. For the testing data, teams will be given a set of new, unlabeled video clips and will need to generate hazard-report captions for each. These outputs will be submitted to the Kaggle platform for scoring. We anticipate concluding the competition with the announcement of winners before the ICCV 2025 workshop so that results can be discussed at the event. Top-performing teams will be invited to present their approaches and findings at the 2COOOL workshop, allowing them to share insights with the research community. As with the previous COOOL competition, the goal isn’t just to top the leaderboard, but to encourage novel approaches that can generalize to detecting and describing hazards in diverse scenarios. By hosting the challenge on Kaggle, we hope to attract a broad range of participants from both academia and industry, tapping into Kaggle’s vibrant community and tools. The impact of the challenge will be measured not only by the accuracy of the winning models, but also by the new techniques and ideas it inspires for vision-language understanding in safety-critical settings.
\section{Conclusion}
\label{sec:conclusion}

The 2COOOL 2025 workshop is poised to advance the conversation on how autonomous vehicles can cope with the unknown unknowns of the road. By gathering researchers working on out-of-label hazard detection, along with practitioners confronting these issues in industry, the workshop creates a vibrant forum for knowledge exchange and collaboration. The invited talks will shed light on state-of-the-art methods and real-world hurdles, while the contributed papers present new algorithms, datasets, and insights that inch us closer to truly safe autonomous driving.

Furthermore, the introduction of a focused challenge on hazard report generation adds an exciting practical dimension, pushing participants to develop systems that not only detect hazards but also explain and contextualize them. Looking ahead, we envision several key takeaways and future directions emerging from 2COOOL 2025. First, improving the robustness and adaptability of perception systems will remain a central goal – the solutions presented here (e.g. open-set recognition techniques, novelty-aware models) could be integrated into next-generation autonomous driving platforms. Second, the role of vision-language models is expected to grow; as demonstrated by our challenge, being able to describe what the AI sees can greatly enhance transparency and safety. This aligns with the broader trend of using AI explainability to build trust in autonomous systems. Third, the workshop underlines the importance of collaboration across domains: lessons from robotics, cognitive science, and even psychology (for understanding human reactions) may inform better hazard response strategies. By continuing to broaden the community tackling these problems, we can approach the goal of covering the long tail of driving scenarios. 

The challenge of out-of-label hazards is a critical frontier for autonomous driving research. Ensuring that self-driving cars can handle rare and unforeseen events is not only a technical endeavor but also a necessity for public safety and acceptance. The 2COOOL workshop – with its blend of research presentations, keynotes, and an innovative competition – contributes to this mission by spotlighting the latest developments and open questions. We are optimistic that the discussions and results from 2COOOL 2025 will spark new ideas and partnerships that accelerate progress toward safer autonomous vehicles. By preparing our models to expect the unexpected, we move closer to a future where autonomous driving can deliver on its promise of enhanced safety and mobility for all. 

Acknowledgments: We thank the 2COOOL program committee, the ICCV 2025 organizers, and our sponsors for their support. Special gratitude goes to all co-organizers of 2COOOL 2025 for their hard work in putting together this workshop. We also appreciate the enthusiasm of the autonomous driving research community – from authors and speakers to challenge participants – for contributing their expertise. Together, we are tackling one of the most challenging aspects of computer vision in transportation, and we look forward to the breakthroughs that will emerge from these collective efforts.

\small
\nocite{*}
\bibliographystyle{unsrt}
\bibliography{ref}

\end{document}